%% file: AAAI_v3.tex
\def\year{2021}\relax
\documentclass[letterpaper]{article} 
\usepackage{aaai21}  
\usepackage{times,bbm}  
\usepackage{helvet} 
\usepackage{courier,color}  
\usepackage[hyphens]{url}  
\usepackage{graphicx} 
\usepackage{mathtools}

\DeclarePairedDelimiter{\floor}{\lfloor}{\rfloor}
\usepackage{natbib}  
\usepackage{caption} 
\usepackage{amsmath,amssymb}
\DeclareMathOperator*{\argmin}{argmin}
\DeclareMathOperator*{\argmax}{argmax}
\usepackage[ruled,vlined]{algorithm2e}
\usepackage{subcaption}
\usepackage{makecell}
\title{A Markov Decision Process Approach to Active Meta Learning}
\author {
        Bingjia Wang,\textsuperscript{\rm 1}
        Alec Koppel, \textsuperscript{\rm 2}
        Vikram Krishnamurthy \textsuperscript{\rm 1} \\
}
\affiliations {
    \textsuperscript{\rm 1} School of Electrical and Computer Engineering, Cornell University \\
    \textsuperscript{\rm 2} CISD, U.S. Army Research Laboratory \\
    bw499@cornell.edu, alec.e.koppel.civ@mail.mil, vikramk@cornell.edu
}

\begin{document}

\maketitle

\begin{abstract}
In supervised learning, we fit a single statistical model to a given data set, assuming that the data is associated with a singular task, which yields well-tuned models for specific use, but does not adapt well to new contexts. By contrast, in meta-learning, the data is associated with numerous tasks, and we seek a model that may perform well on all tasks simultaneously, in pursuit of greater generalization. One challenge in meta-learning is how to exploit relationships between tasks and classes, which is overlooked by commonly used random or cyclic passes through data. In this work, we propose actively selecting samples on which to train by discerning covariates inside and between meta-training sets. Specifically, we cast the problem of selecting a sample from a number of meta-training sets as either a multi-armed bandit or a Markov Decision Process (MDP), depending on how one encapsulates correlation across tasks. We develop scheduling schemes based on Upper Confidence Bound (UCB), Gittins Index and tabular Markov Decision Problems (MDPs) solved with linear programming, where the reward is the scaled statistical accuracy to ensure it is a time-invariant function of state and action. Across a variety of experimental contexts, we observe significant reductions in sample complexity of active selection scheme relative to cyclic or i.i.d. sampling, demonstrating the merit of exploiting covariates in practice.
\end{abstract}

\section{Introduction}
In supervised learning, we learn to map features to targets by minimizing a statistical loss averaged over samples from an unknown distribution which is typically associated with a singular task \cite{miller}. When this map is a universal function approximator, i.e., a deep neural network (DNN), this framework has yielded successes across a variety of applications \cite{yin2017comparative,GOPALAKRISHNAN2017322,7926694,6423452}. However, its successes have been limited when data is comprised of several qualitatively different regimes, or \emph{tasks}. To enhance adaptivity to disparate tasks, meta-learning seeks to obtain model parameters along the Pareto frontier of the minimizer of many training objectives simultaneously \cite{andrychowicz2016learning}, and has gained attention for overcoming data starvation issues in robotics and physical systems \cite{finn2017model}.

Existing approaches, however, offer little guidance about how to select samples on which to train to enable fast convergence, and instead operate via cyclic or random sampling. Doing so is appropriate when disparate tasks are statistically independent. However, in many contexts such as meteorology \cite{NIPS2017_6932}, computer vision, and robotics \cite{finn2017model}, significant relationships between tasks exist. We are then faced with the question of how to incorporate such relationships into the training of a meta-model. In this work, we do so via active sample selection during training meta-models. This active sample selection is executed according to correlation within and across tasks via multi-armed bandits (MAB) \cite{lattimore2020bandit} and Markov Decision Processes (MDPs) \cite{puterman2014markov} based schedulers, which yields substantial gains in sample efficiency across a variety of experimental settings.

Before continuing, a few historical remarks are in order. Augmenting DNN training to improve adaptivity has received substantial interest over the years. Transfer learning relaxes the independent and identically distributed (i.i.d.) hypothesis on data, and seeks to transform a model good for one task to another (domain adaptation) \cite{10.1007/978-3-030-01424-7_27,transfer}, i.e., transfer an understanding of Spanish to Italian \cite{transfer}. Generative modeling, by contrast, directly estimates the data distribution in order to output new examples that plausibly could have been drawn from the original data, similar in spirit to bootstrapping. Recent advances in parameterizing these models using deep neural network, have enabled scalable modeling of complex, high-dimensional data \cite{shorten2019survey}. Both approaches are effective for transferring from one task to another, but it is unclear how to employ these approaches when seeking generalization across many tasks, unless the generative/covariance model co-evolves with data drift, which may cause instability \cite{radford2015unsupervised}.

By contrast, meta-learning seeks to learn attributes of a problem class which are common to many distinct domains, and has been observed to improve adaptability via explicitly optimizing their few-shot generalization across a set of meta-training tasks \cite{wang2019paired}. Importantly, doing so enables learning of a new task with as little as a single example \cite{yu2018one,yin2019metalearning}. Meta-learning algorithms can be framed in terms of a cost that ties together many training sub-tasks simultaneously, with, for instance, recurrent or attention-based models, or an otherwise two-stage objective \cite{liu2019stochastic}:  the inner cost defines performance on a single task, and the outer meta-objective tethers performance across tasks. Doing so results in procedures that experimentally have yielded substantial gains in terms of DNN adaptation and generalization to new tasks \cite{rajeswaran2019metalearning}.

\begin{figure}
    \centering
    \includegraphics[width=0.45\textwidth]{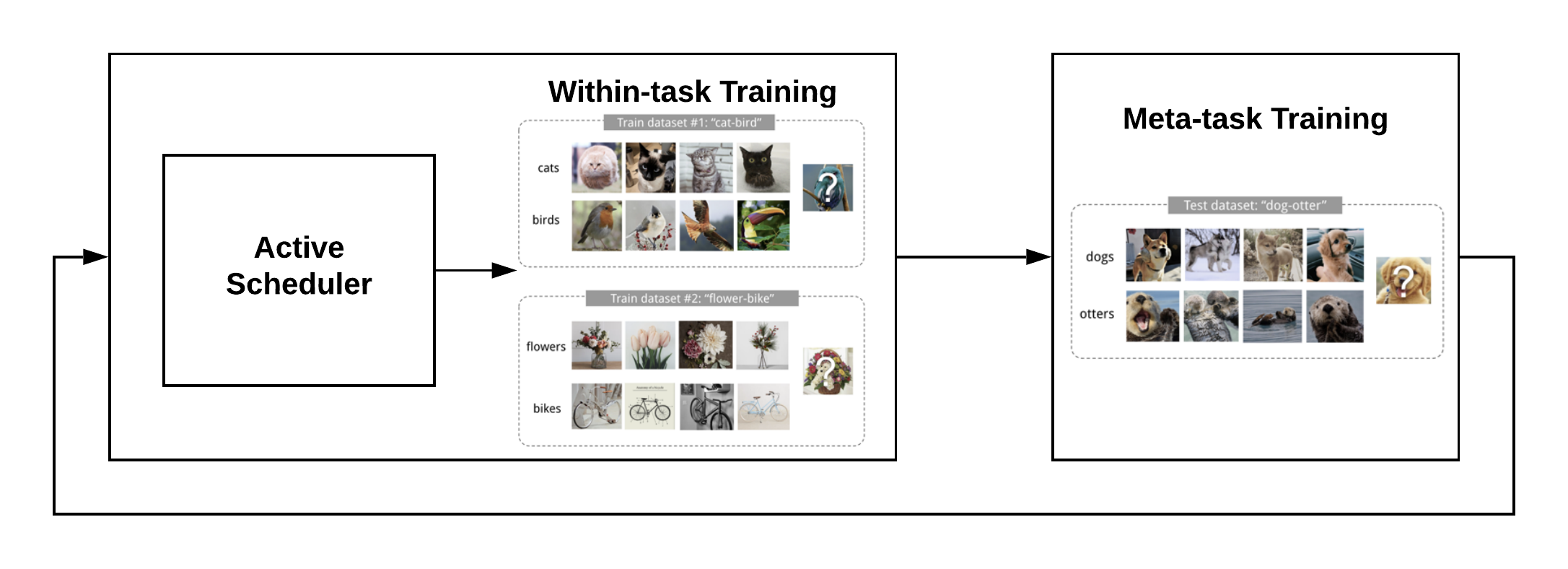}
    \caption{Our scheduler selects which samples from training subsets to execute task-specific updates to ensure the meta-model's performance improves as rapidly as possible as quantified by meta-training subsets' contribution to the meta-model's validation accuracy. Doing so requires a novel definition of the reward in multi-armed bandits or MDPs.}  \vspace{-4mm}
    \label{fig:mesh1}
\end{figure}

The aforementioned works, as well as other meta-learning objectives, operate under the assumption that training samples are i.i.d. to justify sampling cyclically or randomly. This assumption is invalid for settings involving drift or latent relationships between classes, such as training an NLP system for both Spanish and Italian \cite{peters2019tune}, image classification of animals from a common genus \cite{wang2018large}, or systems identification problems arising in ground robotics when traversing prairie and forest floor \cite{koppel2016online,chiuso2019system}. Thus, in this work, we propose to build a scheduler on top of the meta-learner (Figure \ref{fig:mesh1}) to exploit relationships between meta-training data subsets to allocate samples judiciously.

To do so, we incorporate ideas from active learning \cite{cohn1996active}, specifically, selecting a given meta-learning training subset, according to either a multi-armed bandit \cite{cite-key} or a Markov decision process (MDP) \cite{Bel}. Which technique is appropriate depends on whether the statistical accuracy of one task is allowed to be correlated with another. In either case, the state is the weights of a meta-learning model, the arm (action) is the index of the specific training task or class label, and the reward is the statistical accuracy of the meta-model on a validation set multiplied by a scaling factor to ensure the reward is stationary. Moreover, regret of a given arm is the scaled average long-run validation accuracy on that meta-training subset. 

Experimentally, we observe the merit of bandit selections when we employ the Upper Confidence Bound (UCB) or Gittins Index, and MDP policies based upon a linear programming solver \cite{de2003linear} for meta-training DNNs. In particular, we obtain orders of magnitude improvement in sample complexity when employing our sample selection schemes relative to cyclic or random sampling (Table \ref{tab:exp_summary}) for training feedforward multi-layer DNNs and convolutional variants on MNIST \cite{726791}, the real world Extreme Weather dataset \cite{NIPS2017_6932}, and a meta-learning variant of CIFAR100 \cite{cifar}. On top of sample efficiency gains, the order of sample selection experimentally can fundamentally improve the limit points to which the meta-model converges.

\begin{table}[]
    \centering
    \scalebox{0.7}{
    \begin{tabular}[0.45\textwidth]{ |c|c|c|c| } 
    \hline
    & UCB Scheduler & Gittins Index Scheduler & MDP Scheduler \\
    \hline
    Digit Recognition & 24.5 & 32.5 & / \\
    \hline
    Meta CIFAR-100 & 2.5 & 3.57 & / \\
    \hline
    Extreme Weather & 1.25 & 2.42 & 3.33 \\
    \hline
    \end{tabular}
    }\vspace{-2mm}
    \caption{Relative sample efficiency gain compared to baseline cyclic sampling on different experiments.}\vspace{-4mm}
    \label{tab:exp_summary}
\end{table}

\section{Elements of Meta-Learning}
In supervised learning, we seek to build a predictor $f_w: \mathcal{X} \rightarrow \mathcal{Y}$ which maps feature vectors $x\in\mathcal{X}$ to target variables $y\in\mathcal{Y}$ by minimizing a loss function $\ell: \mathbb{R}^{p} \times \mathcal{X} \times \mathcal{Y} \rightarrow \mathbb{R}$ in expectation over the data distribution $\mathbb{P}(x,y)$ which is unknown. Here $w\in\mathbb{R}^p$ denotes the parameters of the statistical model (such as a feedforward or convolutional neural network). The loss $\ell$ quantifies the difference between candidate prediction $f_w(x)$ at an input vector $x\in\mathcal{X}$ and a target variable $y\in\mathcal{Y}$, and is small when $f_{w}(x)$ and $y$ are close. For concreteness and clarity, we focus on the case of multi-class classification, an instance of supervised learning, although the ideas developed in this work are also applicable to unsupervised and reinforcement learning. Thus, the space of target variables is of the form $\mathcal{Y}=\{1,\dots,C\}$, where $C$ is the number of classes. In this context, we wish to compute the parameters that minimize the statistical loss over $w\in \mathbb{R}^p$,
$$w^*=\argmin_{w}\mathbb{E}_{x,y}[\ell(f_{w}(x),y)]$$
where the expectation is over $\mathbb{P}(x,y)$. In practice, one is given a batch of data $\mathcal{D}=\{(x_{1}, y_{1}),...,(x_{k},y_{\tilde{n}})\}$, which may be associated with any number $N$ of unknown distributions $\{\mathbb{P}^i(x,y)\}_{i=1}^N$ colloquially referred to as \emph{tasks}. In particular, we have access to $N$ distinct training subsets $\mathcal{D}^i=\{x_u,y_u\}_{u=1}^{\tilde{n}_i}$ whose union is $\mathcal{D}$, and we would like to find a model that simultaneously performs well on each:
\begin{align}\label{eq:one_stage}
\mathbf{w}^*\!\!=\argmin_{w\in\mathbb{R}^p}  \! \tilde{L}(w)\!:= \! \!\!\!\!\!\!\!\! \!\sum_{\{x_u,y_u\}\in\mathcal{D}^i\!\!\!}\!\!\!\!\!\!\!\!\!\ell(f_{w}(\!x_u),y_u)\! \text{ for } i\!=\!1,\dots,N\!
\end{align}
We consider that each meta-learning sample subset $\mathcal{D}^{i}$ is split into a training and a validation set, i.e., $\mathcal{D}^{i}=\mathcal{D}_{tr}^{i}\cup \mathcal{D}_{val}^{i}$ with $|\mathcal{D}_{tr}^{i}|=n$, and that the training subsets $\mathcal{D}_{\text{tr}}^{i}$ for all $i$ are used for training \emph{within} tasks, whereas the validation set is used \emph{across} tasks. Moreover, we denote $\mathcal{D}_{\text{val}}=\cup_i\mathcal{D}_{\text{val}}^{i}$ and $\mathcal{D}_{tr}=\cup_i\mathcal{D}_{tr}^{i}$. \footnote{For disambiguation, we denote samples of $\mathcal{D}^{i}$ as $\{x_{u}^{i},y_{u}^{i}\}$ for $u=1,\dots,\tilde{n}_i$. Moreover, we denote $n_i$ as the number of training examples available for task $i$. Throughout, to further alleviate notation, we suppress the dependence of example $(x_{u}^{i},y_{u}^{i})$ on class $c$, and instead leave this dependence implicit.} Then, we hypothesize that the statistical model $ f_{w}=f_{w_\lambda}$ depends on a vector of hyperparameters $\lambda\in\mathbb{R}^d$, such as the regularizer, the radius of a pooling step in a convolutional neural network, or other architectural considerations. One way to pose the problem of meta-learning is as a two-stage optimization variant of \eqref{eq:one_stage}:
\begin{align}\label{eq:two_stage}
&\min_{\lambda}L^i(w_{\lambda}):=\!\!\!\!\!\!\!\!\sum_{\{x_u,y_u\}\in\mathcal{D}_{\text{val}}^i}\!\!\!\!\!\!\!\!\!\!\!\ell(f_{w_\lambda}(x_u),y_u) \ \text{ for } i=1,\dots,N \\
&\text{ s.t. } w_{\lambda}\in\argmin_{{w_\lambda}} \!H^i(w_{\lambda}):=\!\!\!\!\!\!\!\!\!\!\sum_{\{x_u,y_u\}\in\mathcal{D}_{\text{tr}}^i}\!\!\!\!\!\!\!\!h(f_{w_\lambda}(x_u),y_u) \nonumber
\end{align}
where $h$ is again some cost, possibly equal to $\ell$, which is small when $f_{w_\lambda}(x_u)$ and $y_u$ are close. This formulation yields models $f_{w_\lambda}$ which both perform well on individual tasks $i$ as quantified by $H^i(w_{\lambda})$ and across tasks through seeking to minimize $L^i(w_{\lambda})$ for all $i=1,\dots,N$ simultaneously. That is, model selection of $f_{w_\lambda}$ according to \eqref{eq:two_stage} at the inner-stage (the constraint evaluation) is decoupled across tasks, whereas at the outer stage, the objective is coupled by hyperparamaters $\lambda$. For connections to bilevel optimization, see \cite{hyper,likhosherstov2020ufo}.

Given that computing the simultaneous minimizer of a number of different non-convex functions is intractable, one may hypothesize that the universal quantifier over task $i$ in \eqref{eq:two_stage} may be replaced by the sum-costs \vspace{-2mm}
\begin{equation}\label{eq:training_cost}
    L(w_{\lambda})=\sum_{i=1}^N L^i(w_{\lambda}) \; , \quad H(w_{\lambda})=\sum_{i=1}^N H^i(w_{\lambda})\; , 
\end{equation}
which presupposes that tasks and classes are statistically \emph{independent}. Then, because exactly solving the inner optimization problem, i.e., the constraint in \eqref{eq:two_stage}, is both intractable numerically when $f_{w_\lambda}$ is a neural network (as the problem becomes non-convex) and may lead to solutions that over-prioritize a singular task (over fit), one may consider the computational approximation of \eqref{eq:two_stage} as \cite{finn2017model}
\begin{align}\label{eq:meta_learning_objective}
\min_{\lambda} L(w_{\lambda}) \quad \text{ s.t. } w_{\lambda} =w_{\lambda} - \eta \nabla_w H(w_{\lambda})\; .
\end{align}
Note that the $argmin$ in the constraint of \eqref{eq:two_stage} been substituted in \eqref{eq:meta_learning_objective} by the fact that we seek model parameters close to the fixed point of the gradient of the task-specific objective $H(w_{\lambda})$ \cite{finn2017model}, while also minimizing the cost $L$ which is defined \emph{across} tasks. The spirit of \eqref{eq:meta_learning_objective} is that we seek model parameters that perform well after a few gradient steps on an unseen task, whereas \eqref{eq:one_stage} yields solutions that perform well on average observing a number of samples from a common distribution. Prevailing practice in meta-learning is built upon assuming statistical independence between tasks and classes, i.e., writing $H=\sum_{i=1}^N H^i$, which permits grouping the inner and outer expectations -- see \cite{fallah2020convergence}. \medskip

{\bf \noindent Main Results}
In this work, we move beyond the hypothesis that tasks and classes are independent by 
 considering a generalization of \eqref{eq:meta_learning_objective}: rather than focusing on the aggregate task-specific cost $H(w_{\lambda})$, we retain the task-specific model fitness in the constraint $H^i(w_{\lambda})$, 
\begin{align}\label{eq:meta_learning_objective2}
\!\!\!\!\!\min_{\lambda} L(w_{\lambda}) \! \ \text{ s.t. } w_{\lambda} \!=w_{\lambda} \!-\! \eta \nabla_w H^i(w_{\lambda}), i=1,\dots,N  ,
\end{align}
which instead reveals the question of how to compute a point at the intersection of a set of $N$ constraints for each of $C$ classes when the satisfaction of one constraint influences another. In this work, we focus on sequential approaches to addressing this question, inspired by active learning \cite{cohn1996active,settles2011theories}. In particular, we develop techniques to select which among the $N$ different tasks and $C$ different classes one should execute a training step at any given time such that the overall meta-learning performance $L(w_{\lambda})$ is optimized expeditiously. Doing so yields significant gains in sample efficiency of training meta-learners across a variety of experimental contexts, as we demonstrate in Sec. \ref{sec:exp} -- see Table \ref{tab:exp_summary}. Next, we shift to the technical development of bandits and MDPs to this end.

\begin{algorithm}[t]
   \SetAlgoLined
   {\bf Initialize:} No. tasks blah $N$, task-specific data $\{\mathcal{D}_{\text{tr}}^i\}$, $|\mathcal{D}^{i}_{tr}|=n$, validation set $\mathcal{D}_{\text{val}}$, init. params. $w_\lambda\in\mathbb{R}^p$ associated w/ hyperparams. $\lambda\in\mathbb{R}^d$, batch size $B$\\
   \For{$k=1,...$}{
   \For{$t=1,...,\floor{\frac{n}{B}}$}{
       Schedule mini-batch $\mathcal{B}(\{\theta_u\})=\{x^{\theta}_{u},y^{\theta}_{u}\}$ \\
       Update parameters $w$ via SGD [cf. \eqref{eq:task_training_step}]\vspace{-2mm}
       %
 $$w_{t+1} \!=w_{t} \!-\! \delta \nabla_w \sum_{u=1}^B h(f_{w_t}(x_u^\theta),y_u^\theta)$$
 \vspace{-3mm}
 %
    }
       Update hyperparams. $\lambda$ of meta-model [cf. \eqref{eq:meta_training_step}]\vspace{-2mm}
       $$ \lambda_{k+1} \!=\lambda_{k} \!-\! \eta  \!\!\!\!\!\!\!\!\!\sum_{\{x_u,y_u\}\subset \mathcal{D}_{val}} \!\!\!\!\!\!\!\!\! \nabla_\lambda\ell(f_{w_N}(x_u),y_u)$$\vspace{-3mm}
       }
    \caption{Active Learning for Meta Learning}
   \label{alg:1}
   \Return{\normalfont Meta-model $f_{w_\lambda}$ params. $w$, hyperparams. $\lambda$} 
\end{algorithm}\setlength{\textfloatsep}{5pt}

\section{Active Sample Selection}

In meta-learning \eqref{eq:meta_learning_objective2}, there are two intertwined challenges. First, to enforce the constraint, one requires access to training examples $(x_u^i,y_u^i)$ for each task $i$ and class $c$ in order to evaluate the gradient of the different task-specific objectives $H^i(w_{\lambda})$ with respect to model parameters $w_{\lambda}$ for fixed hyperparameters $\lambda$. With access to $(x_u^i,y_u^i)$ for each task,  a stochastic gradient update with step-size $\delta>0$ is performed: \vspace{-3mm}
\begin{equation}\label{eq:task_training_step}
 w_{t+1} \!=w_{t} \!-\! \delta \nabla_w \sum_{u=1}^B h(f_{w_t}(x_u^i),y_u^i)\; , \quad
 \end{equation}
where $1\leq B\leq n$ is some mini-batch size, which makes \eqref{eq:task_training_step} a stochastic gradient step (for $B< n$), and we have suppressed dependence on $\lambda$ for succinctness. Existing approaches proceed to execute training steps on all tasks $i$ and classes $c$ cyclically, meaning there are $t=N$ total updates of the form \eqref{eq:task_training_step} -- see \cite{andrychowicz2016learning,finn2017model}. Then, we conduct a stochastic gradient update of step-size $\eta>0$ with respect to the meta-model:
\begin{equation}\label{eq:meta_training_step}
 \lambda_{k+1} \!=\lambda_{k} \!-\! \eta  \!\!\!\!\!\!\!\!\!\sum_{\{x_u,y_u\}\subset \mathcal{D}_{val}} \!\!\!\!\!\!\!\!\! \nabla_\lambda \ell(f_{w_N}(x_u),y_u)\; , \quad
 \end{equation}
 For simplicity, we consider that $B$ samples are chosen from validation set $ \mathcal{D}_{val}$ to execute a meta-model update in \eqref{eq:meta_training_step}.
 
 One way of going beyond statistical independence between tasks in the updates is by using second-order information \cite{im2019model,song2019maml,park2019meta}; however, when computing the Hessian of the Lagrangian of \eqref{eq:meta_learning_objective2}, its statistical properties are only locally (not globally) informative due to non-convexity -- see \cite{nocedal2006numerical}. Instead, we directly exploiting covariates within and between tasks. While related ideas have been proposed for how to weight the gradient of the meta-objective $L(w_\lambda)$ in \cite{cai2020weighted,simon2020modulating,nicholas2020m2sgd}, none have augmented the update rule both \emph{within a task} and across tasks.

To do so, we estimate dependencies both within each task and dependencies across different tasks as respectively a multi-armed bandit (MAB) or a Markov Decision Problem (MDP). Before proceeding to defining their specific use in modeling dependencies to more effectively schedule which task one should perform an inner-loop update at a given time, we present the generic procedure for concreteness as Algorithm \ref{alg:1}, which is depicted graphically in Figure \ref{fig:mesh1}. It involves a MAB/MDP scheduler followed by the within-task and cross-task SGD optimization. Next, we define in detail the Scheduler called in Algorithm \ref{alg:1}.

\begin{figure}[t]
    \centering
    \includegraphics[width=0.35\textwidth]{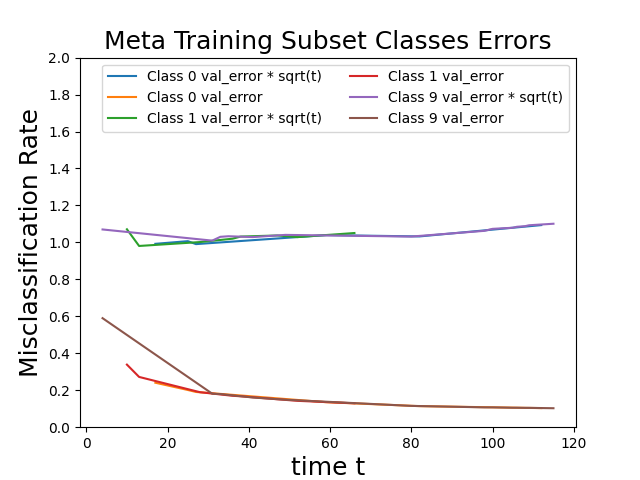}\vspace{-3mm}
    \caption{Scaled $\sqrt{t} \times (\text{validation error})$ on MNIST is nearly constant for each class (state) as a function of within-task training index $t$. Thus, via the approximate relationship between the rate of attenuation of the expected gradient of the meta-training objective $\mathbb{E}[\|\nabla_w L(w_t)\|]$ and validation error $e(t)$ during within-task training, we can define a reward $r(t)= 1 - \sqrt{t} e(t)$ which is time-invariant, and hence satisfies the conditions required for a valid bandit formulation in the sense that the distribution in \eqref{eq:regret} is stationary.}
    \label{fig:reward}
\end{figure}

\subsection{Multi-armed Bandits Scheduling of Subsets}\label{sec:mab}
Multi-armed bandits (MAB) encapsulates the setting where we seek to exploit covariates within a task, e.g., how one class is correlated with another. In MAB, at each time $t$, a player (scheduler) selects one among $S$ available arms, denoted as $\theta_{t}\in\{1,\dots,S\}$ (subsequently we abbreviate $\{1,\dots,S\}:=[S]$), after which a reward $r_{t}(\theta_{t})$ is revealed \cite{lattimore2020bandit}. Since rewards are observed sequentially, under the setting that the underlying generating process of the rewards is stationary, the optimal selection is the one that performs best-in-hindsight, i.e., $\theta^{\ast}=\argmax_{\theta\in\Theta}R(\theta):=\mathbb{E}\{r_{t}(\theta)\}$. The performance of any sequential selection strategy for  $\theta_t$ may be quantified as the expected sub-optimality, or \emph{regret} $R_T$, defined as,\vspace{-2mm}
\begin{equation}\label{eq:regret}\
R_{T}=\mathbb{E}\{T \cdot r_{t}(\theta^{\ast})-\sum_{t=1}^{T}r_{t}(\theta_{t})\} \; .
\end{equation}
Strategies whose time-average regret approaches null, $R_T/T \rightarrow 0$ as the time horizon $T$ becomes large are called no-regret. We consider two widely-used MAB no-regret algorithms, the Upper-Confidence Bound (UCB) \cite{regret,agrawal1995sample,auer2002finite} and Gittins Indices \cite{gittins1979bandit,gittins2011multi}, due to both their simplicity and that they operate upon fairly different principles. Before shifting to describing how $\theta_t$ is selected for these algorithms, we identify how the structural attributes of MABs are well-suited to active sampling for meta-models.

\begin{algorithm}[t]
    \SetAlgoLined
    \caption{UCB Scheduler}
    \KwResult{Batch $\mathcal{B}$}
    {\bf Input:} Time index $t$\;
    {\bf Initialize:}\\
    Upper Bound $U=2$\;
    Exploration factor $\xi>1$\;
    $V_{t,\mathcal{D}^{i}}$: number of visits to subset $\mathcal{D}^{i}_{tr}$ until time t\;
    Use initial model to train on each $\mathcal{D}^{i}$ with first batch of samples $\{x^{i}_{u},y^{i}_{u}\}_{u=1}^{B}$ independently to obtain $r_{0}(\mathcal{D}^{i})$\;
    $V_{0,\mathcal{D}^{i}}=1,\forall i\in[N]$\\
    {\bf At time $t$:}\\
    $\tilde{\mu}_{t-1,\mathcal{D}^{i}} = \frac{1}{V_{t-1,\mathcal{D}^{i}}}\sum\limits_{\tau=0}^{t-1}r_{\tau}(\theta_{\tau})\mathbbm{1}\{\theta_{\tau}=\mathcal{D}^{i}\}, \forall i\in[N]$\\
    $\theta_{t}=\argmax_{\mathcal{D}^{i}}[\tilde{\mu}_{t-1,\mathcal{D}^{i}}+U\sqrt{\frac{\xi\log t}{V_{t-1,\mathcal{D}^{i}}}}]$\\
    $V_{t,\mathcal{D}^{i}}=\sum\limits_{\tau=0}^{t}\mathbbm{1}\{\theta_{\tau}=\mathcal{D}^{i}\},\forall i\in[N]$\\
    $\mathcal{B}=\{x^{\theta_{t}}_{u},y^{\theta_{t}}_{u}\}_{u=(t-1)B+1}^{tB}$
\label{alg:3}
\end{algorithm}

In meta-learning, for multi-class classification with $C^i$ classes for task $i$, the $S$ different possible arms are the $\cup_i [C^i]$ classes, i.e., $[S]=\cup_i [C^i]$, and the arm $\theta_t$ pulled at a given time $t$ is the class $c_t$, meaning that one executes a SGD step \eqref{eq:task_training_step} associated with class $c_t$. An open question is then how to define the reward $r_t(\theta)$. One possibility is the statistical accuracy on the validation set $\mathcal{D}_\text{val}$:\vspace{-1mm}
\begin{equation}\label{eq:validation_accuracy}
\tilde{r}_t(\theta) = \frac{1}{|\mathcal{D}_{\text{val}}|}\sum_{\{x_u,y_u\}\in\mathcal{D}_{\text{val}}} \mathbbm{1}[f_{w_t}(x_u) = y_u] \; ,
\end{equation}
where  the indicator $ \mathbbm{1}[f_{w_t}(x_u) = y_u]$ is $1$ when the model $f_{w_t}$ classifies training example $(x_u,y_u)$ correctly and null otherwise.
Observe, however, that as the model $w$ and hyperparameters $\lambda$ evolve during training, the reward will drift as the validation accuracy improves, which invalidates the stationarity hypothesis (that the distribution in \eqref{eq:regret} is  stationary) underlying the guarantees of UCB and Gittins indices.

To ameliorate this issue, we use the fact that the convergence rate of SGD and its first-order variants (such as Adam) on non-convex problems exhibit a $\mathcal{O}(1/\sqrt{t})$ convergence rate to a first-order stationary point in terms of attenuation of the gradient norm \cite{bottou2018optimization}[Sec. 4.3]. Then, based upon the hypothesis that the rates of attenuation of the gradient norm $\mathbb{E}[\|\nabla_w L(w) \|]$ and the statistical error $e_{t} = 1-\tilde{r}_t(\theta) $ are comparable, $\sqrt{t}e_{t}$ should be constant during training. Thus, we define the reward as 
\begin{equation}\label{eqn:reward}
    r_t(\theta)= 1-\sqrt{t}(1-\tilde{r}_t(\theta))
\end{equation}
Figure \ref{fig:reward} shows the errors of some classes in a sample meta-training subset over the first $120$ training steps in our MNIST experiment (elaborated upon in Section \ref{sec:exp}). Observe that $\sqrt{t}e_{t}$ of each state is approximately a constant over time, which provides evidence to support our hypothesis, and thus substantiates our choice of reward for linking class selection among performance on training subsets $H^i(w)$ with the meta-learning validation objective $L(w)$ [cf. \eqref{eq:meta_learning_objective2}]. The values of $\sqrt{t}e_{t}$ may increase for larger $t$ since the model parameters may settle to the local minima and the error saturates. This is not a problem, however, as later selections influence regret less due to the accumulating sum over time in regret \eqref{eq:regret}. This decrease in importance of later decisions may further be  enforced through discounting that arises in UCB, Gittins Indices, and MDPs as described next.


\begin{algorithm}[tb]
    \SetAlgoLined
    \caption{Gittins Index Scheduler}
    \label{alg:5}
    \KwResult{Batch $\mathcal{B}$}
    {\bf Input:} Time index $t$\;
    {\bf Initilize:}
    Compute Gittins Indices $v^{i}$ of $\mathcal{D}^{i}$ using \textbf{Algorithm \ref{appendix:alg4} in Appendix \ref{appendix:gitt}}\\
    {\bf At time $t$:}\\
        $\theta_{t}=\argmax_{\mathcal{D}^{i}}v^{i}(y^{i}_{(t-1)B+1})$\\
        $\mathcal{B}=\{x^{\theta_{t}}_{u},y^{\theta_{t}}_{u}\}_{u=(t-1)B+1}^{tB}$
\end{algorithm}
\subsubsection{Upper Confidence Bound} \label{ucb}
Upper Confidence Bound (UCB) operates upon the principle of optimism in the face of uncertainty. Specifically, we initialize the model associated with task $i$ via a single iteration of \eqref{eq:task_training_step} on $(x_1^i, y_1^i)$. Then, we count the number of times $\theta=\tilde{\theta}$ has been chosen at time $t$ as $V_{t,c}$ for each $\theta\in[C]$, i.e.,
$V_{t,\tilde{\theta}}=\sum_{\tau=1}^t \mathbbm{1}\{\theta_\tau = \tilde{\theta}  \}$ and its associated average reward:\vspace{-2mm}
$$
\bar{\mu}_{t,\tilde{\theta}} = \frac{1}{V_{t,\tilde{\theta}}} \sum_{\tau=1}^t r_\tau(\theta_\tau) \mathbbm{1}\{\theta_\tau = \tilde{\theta}  \}
$$
Then, UCB selection operates via calibrated perturbation from the sample mean of the reward $\bar{\mu}_b$ as\vspace{-1mm}
$$
\theta_{t+1} =\argmax_{\tilde{\theta}} \bar{\mu}_{\tilde{\theta},t}  + U \sqrt{ \frac{\xi \log t }{V_{t,\tilde{\theta}}}}
$$
where $\xi$ and $U$ are constants that encourage exploration. This procedure is repeated for ${B}-1$ total steps, and achieves regret that is logarithmic in the total number of steps ${B}$, which is precisely the within-task mini-batch size -- see \cite{regret}.
%
%
We set the exploration factor $U = 2$. For each hyperparameter update of $\lambda$, a batch of ${B}$ samples are selected from $\mathcal{D}_{\text{val}}$ according to those classes from $\cup_i [C^i]$ which maximize the upper-confidence bound as determined by Algorithm \ref{alg:3}. Then, these samples are used to update the hyperparameters $\lambda$ w.r.t. the validation loss in \eqref{eq:meta_training_step}.

\subsubsection{Gittins Index}\label{git}
UCB is a frequentist (non-Bayesian) strategy: it does not construct any distributional model for how to select $\theta_t$. Next we consider a Bayesian approach based upon Gittins Index, which may also be shown to be no regret \cite{gittins-theory}. It has the additional merit that it exploits the Markovian dependencies between states by the transition matrix structure. Proceeding with its technical development necessitates a distributional model among states. For task $i$, we construct the count-based measure:
\begin{equation}\label{eq:transition_model}
P^{i}_{c c^\prime}=\frac{\text{number of jumps from label} \ c \text{ to } c^\prime}{\text{number of examples with label} \ c} \; .
\end{equation}
This counting-based construction of the transition matrix between classes in $\mathcal{D}^i_{\text{tr}}$ has precedent in Bayesian filtering \cite{krishnamurthy2016partially}[Ch. 5]. Gittins index is then defined as
%
\begin{equation}\label{eq:gittins_index}
    v^i(\theta)=\max_{\tau >0}\frac{\mathbb{E}^i[\sum_{t=0}^{\tau}\beta^{t}r_t(\theta_{t})|\theta_0=\theta]}{\mathbb{E}[\sum_{t=0}^{\tau}\beta^{t}|\theta_0=\theta]}
\end{equation}
where $\tau$ is a measurable stopping time. Here $v(\theta)$ is called Gittins index associated with reward $r(\theta)$ at state $\theta$, and the expectation $\mathbb{E}^i$ is computed with respect to the distribution $P^{i}_{c c^\prime}$ over labels $[C^i]$ for a fixed $i$. We define the Gittins index identically as \eqref{eq:gittins_index} for each meta-training subset $i$ as $v^i(\theta_i)$. 
\begin{algorithm}[t]
    \SetAlgoLined
    \caption{MDP Scheduler}
    \label{alg:6}
    \KwResult{Batch $\mathcal{B}$}
    {\bf Input:} Time index $t$\;
    {\bf Initilize:}
    Compute Value vectors $V(s)$ solving LP (\ref{eqn:lp})\\
    {\bf At time $t$:}\\
        state $s=(y^{1}_{(t-1)B+1},y^{2}_{(t-1)B+1},...,y^{N}_{(t-1)B+1})$\\
        $a=\argmax_{i\in[N]}[r(s,i)+\sum_{s'}\gamma\mathbb{P}^{i}(s,s')V(s')]$\\
        $\mathcal{B}=\{x^{a}_{u},y^{a}_{u}\}_{u=(t-1)B+1}^{tB}$
\end{algorithm}
 The Gittins Index Theorem establishes that a selection is optimal, i.e., no regret \eqref{eq:regret}, if and only if it always selects an arm with highest Gittins index when there is Markovian dependence on the way label transitions occur \cite{gittins-theory}, with \eqref{eqn:reward} as the reward. To investigate whether this condition holds true, we
 %
%
 use Pearson's chi-squared test to determine whether the evidence supports the examples are not i.i.d. at 95\% confident level (significant level (p-value) of 0.05). Further details and validation of the constructed transition matrices is deferred to Appendix \ref{appendix:chi}. In the experimental settings of Sec \ref{sec:exp}, there is significant evidence that classes exhibit Markovian dependence.
 
 
 Since the reward is a constant for each class (state), based on equation (\ref{eqn:reward}), we approximate the reward $r^{i}_{c}$ of state $c$ in $\mathcal{D}^{i}$ as the accuracy of fitting the first sample of label $c$ in $\mathcal{D}^{i}$ into the initial model. The reward vector of $\mathcal{D}^{i}$ is then $\textbf{r}^{i}=[r^{i}_{1},...,r^{i}_{C}]$. We use largest-remaining-index algorithm \cite{1103989} to compute the Gittins Index of each label in each meta-learning subset $i$ (See Appendix \ref{appendix:gitt}). The Gittins Index Theory \cite{gittins-theory} states that the optimal action is to choose the bandit with highest Gittins Index at each iteration. Gittins indices are computed offline before the actual training process. Gittins Index scheduler is shown in \textbf{Algorithm \ref{alg:5}}. 

\subsection{MDPs for Cross-Correlated Task Scheduling}
In MAB, arms are assumed independent from one another in UCB and Gittins index and correlation \emph{across tasks} is not permitted. However, in many applications of meta-learning, dependencies \emph{across} different training subsets exist. In such a setting, the reward for arm $c$ will not remain frozen when arm $c^\prime$ is chosen. To address this limitation, we consider using MDPs, where transition probabilities and reward functions are defined across subsets (arms) $c$ and $c^\prime$.

An MDP over state space $\mathcal{S}$ and action space $\mathcal{A}$ is one in which, starting from state $s$, and selecting action $a$, one moves to state $s'$ with probability $P_a(s,s')$. Then, a reward $R_a(s,s')$ is revealed. The canonical objective of an MDP is to select actions $\{a_t\}$ so as to maximize the average cumulative return, or value, defined as $v(s)=\mathbb{E}[\sum_{u=0}^H \gamma^u R_a(s,s') \mid s_0 = s ]$,
%
%
where $H\leq \infty$ is the horizon length and $\gamma\in(0,1) $ is a discount factor. It's well-known that the optimal value function satisfies  Bellman's optimality equation \cite{puterman2014markov}:\vspace{-1mm}
\begin{figure}[t]
    \centering
    \includegraphics[width=0.35\textwidth]{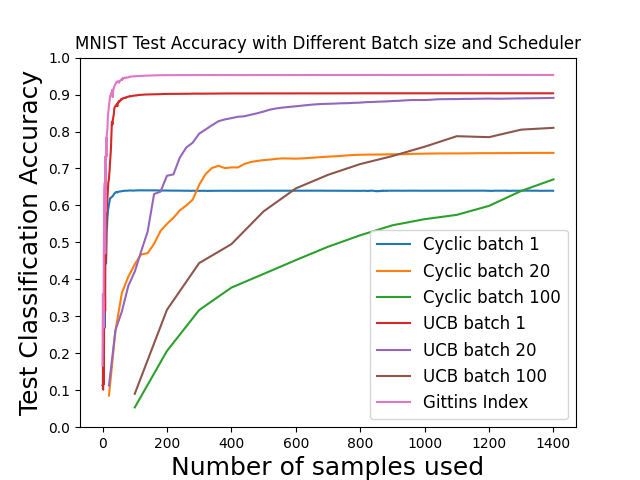}\vspace{-2mm}
    \caption{Digit recognition experiment. Cyclically processing samples from task-specific subsets comprised of Optical Recognition \cite{xu1992methods} and Semeion Handwritten Digits \cite{buscema1998metanet} yields much higher sample complexity for obtaining a well-performing model on unseen MNIST data as compared to bandit schedulers: well-performing models via bandit scheduling only 200 require steps, nearly an order of magnitude reduction. }
    \label{fig:mnist}
\end{figure}
\begin{equation}\label{eq:bellman}
    V(s)=\max_{a}(\sum_{s'}P_{a}(s,s')(R_{a}(s,s')+\gamma V(s')))
\end{equation}
 The optimal policy for each state $s\in\mathcal{S}$ is the action corresponding to the maximum value:\vspace{-2mm}
\begin{equation}\label{eqn:policy}
    a^*=\argmax_{a}(\sum_{s'}P_{a}(s,s')(R_{a}(s,s')+\gamma V(s))) 
\end{equation}
The optimal policy is time-homogeneous, i.e., assigns a fixed action $a$ to any state $s$ independent of time $t$ for $H=\infty$. One way to obtain the optimal policy for tabular settings, i.e., when the state and action spaces are discrete and of moderate cardinality, when the transition matrix is available [cf. \eqref{eq:transition_model}] is via linear programming (LP) \cite{de2003linear}. 
%
%
\begin{figure}[t]
    \centering
    \includegraphics[width=0.35\textwidth]{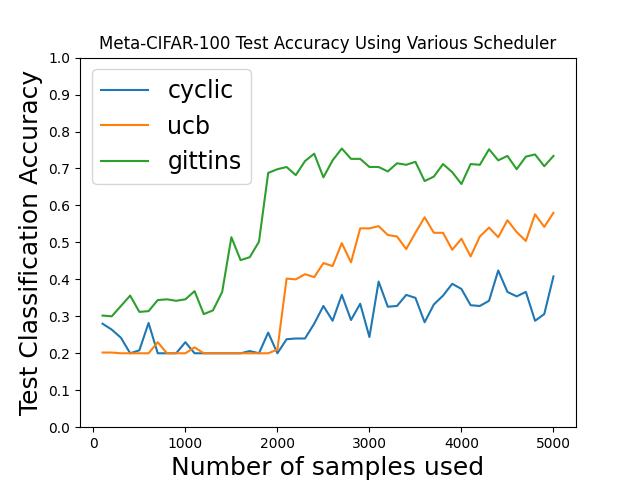}\vspace{-2mm}
    \caption{Meta-CIFAR-100 experiment. CIFAR-100 is divided into task-specific datasets by superclasses "aquatic mammals", "medium-sized mammals", "small mammals" and "insect." Then, we use the superclass  "large carnivores" as the cross-task test set. The performance gap between cyclic and active sampling is more stark for this setting, as the inherent correlation is more pronounced. Gittins Index scheduler achieves 73\% accuracy and UCB achieves 58\% accuracy, while cyclic sampling only has 40\% accuracy.}
    \label{fig:cifar}
\end{figure}
We proceed to formulate this LP for the meta-learning scheduler policy. The state space $\mathcal{S}$ is vector-valued consisting of the $N$-fold Cartesian product of the set of classes $[C]\times \cdots [C]$, the aggregate transition model is the $N$-fold Kronecker product of task-specific transition matrix \eqref{eq:transition_model}, i.e., $\mathbb{P}^{i}=P^{i}\otimes I^{1} \cdots\otimes I^{N-1}$. The Kronecker product ensures the dimensionality consistence between state space $\mathcal{S}$ and the transition model $\mathbb{P}^{i}$.  The action determines which meta-training subset should be chosen at the next training time-slots. 
Moreover, the reward is given as the validation accuracy \eqref{eqn:reward}, as in the beginning of Sec. \ref{sec:mab}, except now we reinterpret the reward as being not only a function of the selected class but also the meta-learning subset $\mathcal{D}^i$ as well, i.e., $r(\theta)=r(s,i)$. This is the additional expressive power of MDPs over Gittins Index. In MDPs, the reward for the same state changes when different arms are played, which exploits both within \emph{and} cross-task correlation. 
%
Then, we formulate an LP to solve for the optimal value $V(s)$:
\begin{align}\label{eqn:lp}
   \!\! \min\sum_{s}\!V(s)
    \text{, s.t.}V(s)\!\geq r(s,i)+\!\sum_{s'}\!\gamma\mathbb{P}^{i}(s,s')V(s')
\end{align}
$\text{ for }\forall s,i \nonumber$.
The optimal policy is computed by equation \eqref{eqn:policy}, where $V(s')$ is obtained from the optimal solution in LP (\ref{eqn:lp}). The MDP scheduler is shown in Algorithm \ref{alg:6}. With our various active selection schemes defined, we shift to establishing their experimental merits for improving the training of meta-models across a variety of problem contexts. 

\section{Experiment}\label{sec:exp}\vspace{1mm}
We experiment the proposed MAB/MDP scheduler on three datasets with either explicit or inexplicit sample dependencies within and cross tasks. Across all experiments, we observe significant relative sample efficiency gain compared to basic cyclic sampling, demonstrating the merit of exploiting covariates in practice.

\subsubsection{Digit Recognition}\label{mnist}
We first evaluate the performance of the schedulers on MNIST handwritten digits \cite{lecun1998mnist} -- MNIST forms the validation set $\mathcal{D}_{\text{val}}$, and the task-specific subsets are the related Optical Recognition \cite{xu1992methods} and Semeion Handwritten Digit data sets \cite{buscema1998metanet} -- see Appendix \ref{appendix:exp_setup} for additional details. 

In cross-task $L_{w_\lambda}$, We select multinomial logistic as the loss $l$, and in task specific $H^{i}(w_\lambda)$,  cross-entropy is selected as lss $l^i$ \cite{murphy2012machine}. The specific model $f_{w_\lambda}$ is a four-layer fully-connected neural network with 300 nodes per layer, and the hyperparameters $\lambda$ concatenates the inner objective's (the constraint in \eqref{eq:meta_learning_objective2}) learning rate and the initialization $w^{i}$. We use Adam \cite{kingma2014adam} with decaying learning rate as outer objective optimizer.

To evaluate the performance, we vary the batch size $B\in\{1,20,100\}$. We compare UCB (Algorithm \ref{alg:3}), Gittins Index (Algorithm \ref{alg:5}), and cyclic sampling from all subsets, where one simply passes through rows of training data one after another. Results are given in Figure \ref{fig:mnist}. 
%
Because there are no strong inner dependencies between examples in MNIST dataset, Gittins index algorithm does not exhibit significant gains compared to UCB. However, both active schedulers outperform the cyclic sampling: to obtain test accuracy 80\%, Gittins index requires 40 samples as compared with 53 for UCB sampling and 1300 for cyclic from test data.

\begin{table}[t]
    \centering
    \scalebox{0.7}{
    \begin{tabular}[0.45\textwidth]{ |c|c|c|c|c|c|c| } 
    \hline
        & UBOT & TMQ & U850 & V850 & VBOT & Z100 \\
    \hline
    MDP & 0.901 & 0.873 & 0.917 & 0.870 & 0.774 & 0.842   \\
    \hline
    Gittins Index & 0.904 & 0.836 & 0.845 & 0.653 & 0.738 & 0.877 \\
    \hline
    UCB & 0.673 & 0.649 & 0.684 & 0.421 & 0.600 & 0.619 \\
    \hline
    Cyclic & 0.352 & 0.043 & 0.304 & 0.480 & 0.592 & 0.448 \\
    \hline
    \end{tabular}
    } \vspace{-2mm}
    \caption{Overall Test Classification Accuracy on Various Features using Different Schedulers. MDP and Gittins Index Schedulers outperform UCB and cyclic scheduling.}\vspace{-1mm}
    \label{tab:acc_sum}
\end{table}

\subsubsection{Meta-CIFAR-100}
The CIFAR-100 dataset is an image dataset containing 100 classes with 600 images each \cite{krizhevsky2009learning}. We construct 4 task-specific meta-training subsets: each task is associated with a superclass, that is, we form meta-training subsets consisting entirely of a single superclass. This defines a classification problem associated with those classes within it -- see Appendix \ref{appendix:exp_setup}.

We use cross entropy as both the inner and outer loss functions and employ a four-layer CNNs with strided convolutions and 64 filters per layer. The hyperparameters are the same as in the Digit Recognition -- see Appendix \ref{appendix:exp_setup}.

Figure \ref{fig:cifar} shows the result of using Gittins Index and UCB compared with cyclic sampling. Note the significant improvements in sample efficiency and the superior limit point to which the model converges when using active selection as compared with cyclic passes through task-specific samples. Moreover, Gittins index outperforms UCB, which is evidence that inherent correlation in the class and task structure is more pronounced for this setting. To achieve 40\% accuracy, Gittins Index scheduler requires 1400 samples, while UCB requires 2000 samples and cyclic scheduler needs 5000 samples, meaning they are respectively $2.57 \times$ and $1.50 \times $ more efficient than cyclic sampling.

\begin{figure}[t]\hspace{-5mm}
    \centering
    \begin{subfigure}[b]{0.25\textwidth}
        \includegraphics[width=\textwidth]{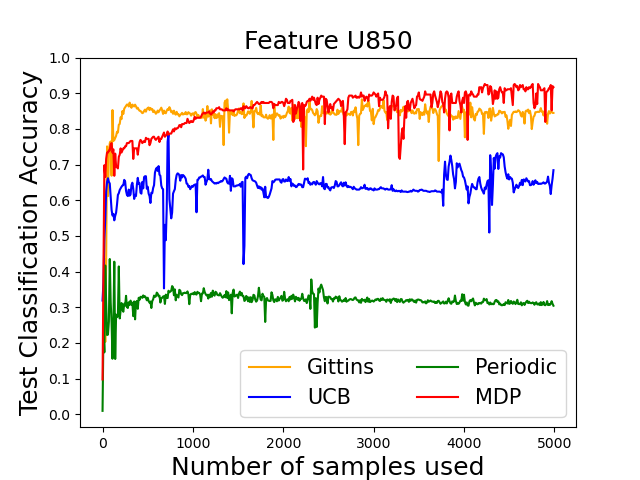}
        \caption{U850}
    \end{subfigure}
    \begin{subfigure}[b]{0.25\textwidth}\hspace{-5mm}
        \includegraphics[width=\textwidth]{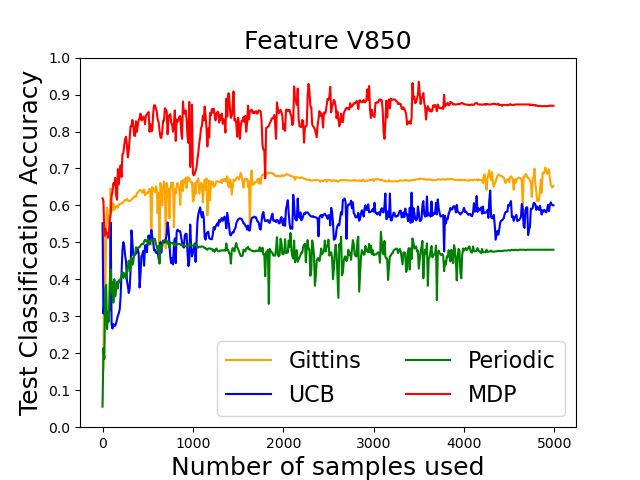}
        \caption{V850}
    \end{subfigure}
    \caption{Evolution of multi-Classification accuracy when using various features. MDP and Gittins Index Schedulers outperform UCB and cyclic scheduling.}
    \label{fig:climate MDP}
\end{figure}

\subsubsection{Extreme Weather}
Gittins index, as compared to UCB, employs the Markovian transition matrix [cf. \eqref{eq:transition_model}] to select the next sample \eqref{eq:gittins_index}, and thus leverages dependencies between classes. In principle, the merit of modeling correlations may be greater when the order of the data has physical meaning. This is not obvious in the case for Meta CIFAR-100 and Digit Recognition. To further investigate the merit of exploiting covariates between samples, we focus on an instance arising in meteorology, as the physical meaning of ordering is inherent due to, e.g., the water cycle.

{\bf \noindent Data Preparation}
We consider the Extreme Weather Dataset \cite{NIPS2017_6932}: training data consists of image patterns of various features and the bounding boxes (prescribed regions) on the images label a specific extreme weather type (considered as class). We use various bounding boxes with different features to construct the meta training, validation and test sets -- see Appendix \ref{appendix:exp_setup} for details.



{\bf \noindent Result}
Our results are summarized in Table \ref{tab:acc_sum} and Figure \ref{fig:climate MDP}. In Appendix \ref{appendix:weather}, one may observe that the constructed transition matrices are diagonally dominant, meaning that covariates between neighboring events/classes are more significant. Thus, it is no surprise that in Table \ref{tab:acc_sum}, one may observe that MDP and Gittins index schedulers outperform other two scheduling policies in all experiments, as they are designed to exploit correlation. Mostly, MDP outperforms Gittins Index, showing that cross-task covariates also have obvious positive effect during training; whereas in some cases, UCB performs comparably to periodic sampling. 

We also compare our results with \cite{DBLP:journals/corr/LiuRPCKLKWC16}, which uses a CNN with hyperparameter optimization to perform the binary classifications on different weather events using multiple features. We use similar features as \cite{DBLP:journals/corr/LiuRPCKLKWC16} described, but with single feature in each test. Although the accuracy we obtain is not comparable, we obtain moderate accuracy with a much simpler correlation model. Specifically, with only 5000 five-features images of size of 32 $\times$ 32, which is 90\% fewer examples than \cite{DBLP:journals/corr/LiuRPCKLKWC16}, we achieve 70-90\% of the accuracy. Moreover, we focus on multi-class problems, which is significantly more challenging than binary classification. Thus, MDPs and Gittins Index schedulers can significantly improve training efficiency. See Appendix \ref{appendix:weather} for further details.

\section{Conclusion}

We departed from prior works on meta-learning that presume independence between tasks by directly considering within and across-task correlation. We proposed a module to select samples according to their contribution to meta-model validation accuracy, which yielded significant sample efficiency gains across a variety of domains as compared to cyclic passes through data. Rigorously analyzing these sample efficiency gains is the subject of future work.

\bibliography{reference}

\input{Appendix}

\end{document}

%% file: Appendix.tex
\appendix
\newpage\onecolumn
\section*{\centering Supplementary Material for \\``A Markov Decision Process Approach to Active Meta Learning"}
In the supplementary material, we provide additional details regarding the construction of meta-learning tasks and evaluations, the associated data sets, and quantities constructed toward these ends. 
\section{Determine Sample Dependencies in Meta-training Subsets Using Chi-squared Test}\label{appendix:chi}
First, we focus on the statistical validation of the transition matrices constructed as \eqref{eq:transition_model} for the various data sets. These transition matrices are essential to the constructing Gittins Index \eqref{eq:gittins_index} and the policy associated with an MDPs \eqref{eqn:lp}. Our goal here is to determine whether the constructed transition matrices provide evidence that classes and tasks exhibit any significant correlation effects.

To do so, we use the Pearson's Chi-Squared to determine whether there is a statistically significant difference between the expected frequencies and the observed frequencies at the 95\% confident level, i.e., p value of 0.05.  The null hypothesis is samples are i.i.d. in each subset. If the statistical test rejects the null hypothesis, i.e., p-value $\leq$ 0.05, Gittins Index or MDPs are justified for scheduling. Under independence, the rows of the constructed Markov chain induced by the transition matrix $P^{i}_{c c^\prime}$ are identical for a fixed $\mathcal{D}_{\text{tr}}^{i}$. Table \ref{tab:p-value} shows the p-values of meta-training subsets in MNIST and meta CIFAR-100 experiments. The p-values of subsets in Extreme Weather experiment are all nearly 0. 

\begin{table}[h]
    \centering
    \begin{subtable}{0.7\textwidth}
    \begin{tabular}{|c|c|c|c|c|c|}
        \hline
        &  Subset 1 & Subset 2 & Subset 3 & Subset 4 & Subset 5\\
        \hline
        p-value & $4.36\times10^{-7}$ & 0.0314 & 0.00836 & $2.33\times10^{-6}$ & $1.20\times10^{-5}$ \\
        \hline
    \end{tabular}\vspace{-1mm}
    \caption{Digit Subsets}\vspace{2mm}
    \end{subtable}
    
    \begin{subtable}{0.5\textwidth}
    \begin{tabular}{|c|c|c|c|c|}
        \hline
        &  Subset 1 & Subset 2 & Subset 3 & Subset 4\\
        \hline
        p-value & 0.0302 & 0.00986 & 0.00215 & 0.00351 \\
        \hline
    \end{tabular}\vspace{-1mm}
    \caption{Meta CIFAR-100}
    \end{subtable}
    
    \caption{p-values of meta-training subsets in MNIST and Meta CIFAR-100. p-values for the Extreme Weather data set are identically near null, and the transition matrix is diagonally dominant -- see Appendix \ref{appendix:weather}.}
    \label{tab:p-value}
\end{table}

\begin{algorithm}[h]
    \SetAlgoLined
    \KwResult{Gittins Indices $v^{i}$}
    State (label) space $\mathcal{Y}=\{1,...,C\}$\\
    N meta training subsets $\{\mathcal{D}^{i}\}_{i=1}^{N}$, $\mathcal{D}^{i}=\{x_{u},y_{u}\}_{u=1}^{\tilde{n}}$ \\
    Transition Matrices of each subset $P^{i}$, $i=1,...,N$ \\
    Discount factor $\beta$ \\
    \For{$i=1,...,N$}{
        Fit the first sample of each label in $\mathcal{D}^{i}$ into the initial model independently and get the reward vector $\textbf{r}^{i}=[r^{i}_{1},...,r^{i}_{C}]$
    }
    \For{$i=1,...,N$}{
        Compute gittins index $v^{i}$ of each subset $\mathcal{D}^{i}$:\\
        Initialization: \\
        state $\alpha_{1}=\argmax\limits_{\alpha \in \mathcal{Y}}r^{i}_{\alpha}$ \\
        $v^{i}(\alpha_{1})=r^{i}_{\alpha_{1}}$ \\
        \For{$l=2,...,C$}{
            $\mathcal{C}(\alpha_{l})=\{\alpha_{1},...,\alpha_{l-1}\} \; , \ 
            \mathcal{S}(\alpha_{l})=\mathcal{Y} \backslash C(\alpha_{l})$ \\
            $Q_{a,b}^{l}=\left\{\begin{array}{rcl} P^{i}_{ab} & \mbox{for} \; b\in \mathcal{C}(\alpha_{l}) \\ 0 & \mbox{otherwise} \end{array}\right. \forall a,b\in\mathcal{Y}$ \\
            $d^{(l)}=[I-\beta Q^{(l)}]^{-1}\textbf{r}^{i} \; , \
            b^{(l)}=[I-\beta Q^{(l)}]^{-1}\mathbbm{1}$ \\
            choose $\alpha_{l}=\argmax\limits_{\alpha \in \mathcal{S}(\alpha_{l})} \frac{d_{\alpha}^{(l)}}{b_{\alpha}^{(l)}}$\\
            $v^{i}(\alpha_{l})=\frac{d_{\alpha_{l}}^{(l)}}{b_{\alpha_{l}}^{(l)}}$ 
        }
    }
    \caption{Compute Gittins Indices of States in Meta Training Subsets}
    \label{appendix:alg4}
\end{algorithm}
This provides substantial evidence across the different data domains that classes and tasks exhibit Markovian dependence, which is evidence that exploiting correlation effects may be useful for scheduling.

\section{Largest-remaining-index Algorithm for Gittins Index in Meta Learning}\label{appendix:gitt}
We use largest-remaining-index algorithm to compute the Gittins Index of each state (class) in each meta-learning subset $i$. We elaborate upon how this procedure works next. Suppose the state space for a given subset is $\mathcal{Y}=\{1,...,C\}$. First step is to identify state (class) $\alpha_{1}$ with the highest Gittins index:
$$
\alpha_{1}=\argmax_{\alpha \in \mathcal{Y}}r^{i}_{\alpha}, 
v^{i}(\alpha_{1})=r^{i}_{\alpha_{1}}
$$
Next step is the recursion to find state $\alpha_{l}$ with $l$th largest Gittins index. Define continuation set as $\mathcal{C}(\alpha_{l})=\{\alpha_{1},...,\alpha_{l-1}\}$ and stopping set as $\mathcal{S}(\alpha_{l})=\mathcal{Y} \backslash \mathcal{C}(\alpha_{l})$. Then state $\alpha_{l}$ and its associated Gittins Index can be computed using a matrix $Q\in\mathbb{R}^{C\times C}$ and two vectors $\textbf{d,b}\in\mathbb{R}^{C}$, which are shown in detail in \textbf{Algorithm \ref{appendix:alg4}}. 
This procedure is then used in the Gittins Index based scheduler summarized in Algorithm \ref{alg:5}.

\section{Additional Details of Experiments}\label{appendix:exp_setup}

We elaborate upon the meta-learning problem formulation in terms of data preparation and allocation, parameter selection, loss function specification, etc. for the experimental results presented in Section \ref{sec:exp}. These points are collated into  Table \ref{appendix:tab:setup} for convenience.

\begin{table}[h]
    \centering
    \scalebox{0.64}{
    \begin{tabular}{|c|c|c|c|c|c|}
    \hline
    &  Meta-training subsets & Within-task loss $h$ & Cross-task loss $f$ & Neural net & Hyperparameters\\
    \hline
    Digit Recognition & \makecell{2 subsets from Semeion Dataset \\ 3 subsets from Opt. Reconition Dataset \\ 1400 samples each subset} & Cross-entropy & Multinomial logistic & \makecell{4-layer fully connected DNN \\ 300 nodes per layer} & \makecell{DNN initial weights $w^i$ and biases $b^i$ \\ Within-task objective learning rate} \\
    \hline
    Meta CIFAR-100 & \makecell{4 subsets from superclasses \\ aquatic mammals, medium-sized mammals \\ small mammals, insect \\ 500 samples per subset} & Cross-entropy & Cross-entropy & \makecell{4-layer CNNs with strided convolutions \\ 64 filters per layer} & \makecell{DNN initial weights $w^i$ and biases $b^i$ \\ Within-task objective learning rate} \\
    \hline
    Extreme Weather & \makecell{5 subsets from first 5 bounding boxes \\ each subset conatains different 5 features \\ 500 samples per subset} & Cross-entropy & Cross-entropy & \makecell{4-layer CNNs with strided convolutions \\ 64 filters per layer} & \makecell{DNN initial weights $w^i$ and biases $b^i$ \\ Within-task objective learning rate} \\
    \hline
    \end{tabular}
    }
    \caption{Experimental setup: data description, parameter selection, architecture specification, loss functions, meta-model definition.}
    \label{appendix:tab:setup}
\end{table}

\subsection{Digit Recognition}
We construct $N=5$ meta-training subsets with $1400$ samples per set. Two are selected from Semeion dataset, and the data from the other three sets are from Optical Recognition Dataset. We construct a common validation set with size 1400 from the two datasets above to evaluate the performance after each hyper iteration. The performance of this procedure is evaluated on a test set comprised of 60000 samples from MNIST dataset. The size of digit images from Optical Recognition dataset and Semeion dataset is different from the size of MNIST images. So we resize the traning and validation image to $28\times28$ in order to ensure images have compatible dimensionality.

\subsection{Meta CIFAR-100}
The CIFAR-100 dataset is an image dataset containing 100 classes with 600 images each \cite{krizhevsky2009learning}. There are 500 training images and 100 testing images per class. The 100 classes are grouped into 20 superclasses, each of which contains classes. Each image comes with a ``fine" label (the class to which it belongs) and a ``coarse" label (the superclass to which it belongs). We construct the task-specific subsets where each task is associated with a superclass, that is, we form data sets consisting entirely of a single superclass, which defines a classification problem associated with those classes within it. Superclasses consist of ``aquatic mammals", ``medium-sized mammals", ``small mammals" and ``insect." Then, we use the superclass  ``large carnivores" as the cross-task validation set. This construction we call Meta-CIFAR-100.

\subsection{Extreme Weather}
We consider the Extreme Weather Dataset \cite{NIPS2017_6932}, where samples from both climate simulations and re-analysis are considered. The reanalysis samples are generated by assimilating observations into a climate model. Ground truth labeling of various events is obtained via multivariate threshold based criteria implemented in TECA, and manual labeling by experts \cite{NIPS2017_6932}. Training data consists of image patterns, where several relevant spatial variables are stacked together over a prescribed region (called bounding box) that bounds a type of weather event, which is considered as ground truth label. The dimension of the bounding box is based domain knowledge of events observed in the real word. There are 1460 example images (4 per day, 365 days in the year) arranged in time order for each year's dataset. We only used 2005's dataset for the experiment. Each image has 16 channels corresponding to 16 features. Each channel is 768 x 1152 corresponding to one measurement per 25 square km on earth. 

We first build the Meta training subsets. For each image, there are up to 15 bounding boxes, where each box indicates a prescribed region in the image that bounds a type of extreme weather event. We used these bounding boxes to split the dataset into different subsets of meta-training set. The first box of each image forms the first subset, the second boxes form the second subset, and so on. Only the first 5 boxes of each image are used, so in total we have 5 different tasks. In order to better differentiate tasks, each subset uses different 5 among 16 features and the features used in each subset are not identical. The first five bounding boxes forms the 5 subsets with 500 images each, another 50 images with all bounding boxes and 5 features are used for validation and other images with all bounding boxes with only one feature are used for testing. Because of the spatial dimension of climate events vary significantly and the spatial resolution of source data is non-uniform, the bounding boxes are resized to 32 $\times$ 32.

\section{Additional Result of Extreme Weather Experiment} \label{appendix:weather}
We present a sample transition matrix of the task-specific data subset via \eqref{eq:transition_model} below:
\begin{equation*}
    \begin{bmatrix}
    0.721 & 0.256 & 0.020 & 0.003\\
    0.052 & 0.901 & 0.033 & 0.014\\
    0.004 & 0.037 & 0.939 & 0.020\\
    0.000 & 0.017 & 0.454 & 0.529
\end{bmatrix}
\end{equation*}
The transition matrix is diagonal-dominant which means that the examples in the dataset are highly correlated. The same type of weather event or its neighbor type of event are likely to happen after one type of extreme weather happens. Combining this structure of likelihood with reward vectors obtained, which are the initial validation accuracy, the Gittins Index reflects the relative "importance" of each state in each arm during the training process. Following the Gittins Index policy we can find the optimal stopping time on one meta-training set and the next dataset the ML model should learn.

Table \ref{tab:feature9sum} displays the summary of examples used in each meta training subset to train the ML model using different schedulers, and feature U850 in test set. Observe that for MDP and Gittins Index scheduler, each meta-training subset contributes to training different types of weather events while training set 4 is rarely scheduled, which indicates that it contributes little towards validation performance for any of type of events. This filtering out of irrelevant information makes training the meta-learner more efficient. The overall classification accuracy for each weather type at the end of training is summarized in Table \ref{tab:feature9acc}. Since the schedulers select more samples labeled as Tropical Cyclone and Extratropic Cyclone, the classification accuracy on these weather types are higher in general. 

\begin{table}[h]\vspace{-2mm}
    \centering
    \begin{subtable}{0.5\textwidth}
    \scalebox{0.7}{
    \begin{tabular}{ |c|c|c|c|c| } 
    \hline
    & Trop. Depression & Trop. Cyclone & Extratropic Cyclone & Atmo. River \\ 
    \hline
    Subset 1 & 140 & 0 & 0 & 0 \\ 
    \hline
    Subset 2 & 10 & 3190 & 0 & 0 \\
    \hline
    Subset 3 & 230 & 0 & 1150 & 0 \\
    \hline
    Subset 4 & 0 & 0 & 20 & 0 \\
    \hline
    Subset 5 & 0 & 0 & 0 & 260 \\
    \hline
    \end{tabular}
    }
    \caption{MDP Scheduler} \vspace{2mm}
    \end{subtable}
    \begin{subtable}{0.5\textwidth}
    \scalebox{0.7}{
    \begin{tabular}[0.4\textwidth]{ |c|c|c|c|c| } 
    \hline
    & Trop. Depression & Trop. Cyclone & Extratropic Cyclone & Atmo. River \\ 
    \hline
    Subset 1 & 440 & 20 & 0 & 0 \\ 
    \hline
    Subset 2 & 0 & 1870 & 0 & 0 \\
    \hline
    Subset 3 & 0 & 10 & 2450 & 0 \\
    \hline
    Subset 4 & 0 & 10 & 0 & 0 \\
    \hline
    Subset 5 & 0 & 0 & 0 & 210 \\
    \hline
    \end{tabular}
    }
    \caption{Gittins Index Scheduler} \vspace{2mm}
    \end{subtable}
    \begin{subtable}{0.5\textwidth}
    \scalebox{0.7}{
    \begin{tabular}[0.4\textwidth]{ |c|c|c|c|c| } 
    \hline
    & Trop. Depression & Trop. Cyclone & Extratropic Cyclone & Atmo. River \\ 
    \hline
    Subset 1 & 120 & 830 & 50 & 0 \\ 
    \hline
    Subset 2 & 170 & 650 & 180 & 0 \\
    \hline
    Subset 3 & 90 & 430 & 490 & 0 \\
    \hline
    Subset 4 & 90 & 140 & 670 & 100 \\
    \hline
    Subset 5 & 40 & 90 & 700 & 160 \\
    \hline
    \end{tabular}
    }
    \caption{UCB Scheduler}
    \end{subtable}\vspace{-2mm}
    \caption{Summary of Examples used in Meta-training subsets, each subset uses different 5 features. The test set uses feature U850. By exploiting correlation, samples associated with certain classes and tasks are significantly down-sampled.}
    \label{tab:feature9sum}
    \vspace{-3mm}
\end{table}

\begin{table}[h]\vspace{-3mm}
    \centering
    \scalebox{0.7}{
    \begin{tabular}[0.45\textwidth]{ |c|c|c|c|c| } 
    \hline
    & Trop. Depression & Trop. Cyclone & Extratropic Cyclone & Atmo. River \\
    \hline
    MDP & 0.789 & 0.961 & 0.947 & 0.658  \\
    \hline
    Gittins Index & 0.421 & 0.836 & 0.963 & 0.395  \\
    \hline
    UCB & 0.368 & 0.698 & 0.788 & 0.421  \\
    \hline
    \end{tabular}
    }
    \caption{Test Classification Accuracy of each Weather Type using Feature U850}
    \label{tab:feature9acc}
\end{table}

%% file: AAAI_v3.bbl
\begin{thebibliography}{56}
\providecommand{\natexlab}[1]{#1}
\providecommand{\url}[1]{\texttt{#1}}
\providecommand{\urlprefix}{URL }
\expandafter\ifx\csname urlstyle\endcsname\relax
  \providecommand{\doi}[1]{doi:\discretionary{}{}{}#1}\else
  \providecommand{\doi}{doi:\discretionary{}{}{}\begingroup
  \urlstyle{rm}\Url}\fi

\bibitem[{Agrawal(1995)}]{agrawal1995sample}
Agrawal, R. 1995.
\newblock Sample mean based index policies with O (log n) regret for the
  multi-armed bandit problem.
\newblock \emph{Advances in Applied Probability} 1054--1078.

\bibitem[{Andrychowicz et~al.(2016)Andrychowicz, Denil, Gomez, Hoffman, Pfau,
  Schaul, Shillingford, and De~Freitas}]{andrychowicz2016learning}
Andrychowicz, M.; Denil, M.; Gomez, S.; Hoffman, M.~W.; Pfau, D.; Schaul, T.;
  Shillingford, B.; and De~Freitas, N. 2016.
\newblock Learning to learn by gradient descent by gradient descent.
\newblock In \emph{Advances in neural information processing systems},
  3981--3989.

\bibitem[{Auer, Cesa-Bianchi, and Fischer(2002{\natexlab{a}})}]{cite-key}
Auer, P.; Cesa-Bianchi, N.; and Fischer, P. 2002{\natexlab{a}}.
\newblock Finite-time Analysis of the Multiarmed Bandit Problem.
\newblock \emph{Machine Learning} 47(2): 235--256.
\newblock \doi{10.1023/A:1013689704352}.
\newblock \urlprefix\url{https://doi.org/10.1023/A:1013689704352}.

\bibitem[{Auer, Cesa-Bianchi, and Fischer(2002{\natexlab{b}})}]{auer2002finite}
Auer, P.; Cesa-Bianchi, N.; and Fischer, P. 2002{\natexlab{b}}.
\newblock Finite-time analysis of the multiarmed bandit problem.
\newblock \emph{Machine learning} 47(2-3): 235--256.

\bibitem[{Bellman(1957)}]{Bel}
Bellman, R. 1957.
\newblock A Markovian Decision Process.
\newblock \emph{Indiana Univ. Math. J.} 6: 679--684.
\newblock ISSN 0022-2518.

\bibitem[{Bottou, Curtis, and Nocedal(2018)}]{bottou2018optimization}
Bottou, L.; Curtis, F.~E.; and Nocedal, J. 2018.
\newblock Optimization methods for large-scale machine learning.
\newblock \emph{Siam Review} 60(2): 223--311.

\bibitem[{Buscema(1998)}]{buscema1998metanet}
Buscema, M. 1998.
\newblock Metanet*: The theory of independent judges.
\newblock \emph{Substance use \& misuse} 33(2): 439--461.

\bibitem[{Cai et~al.(2020)Cai, Sheth, Mackey, and Fusi}]{cai2020weighted}
Cai, D.; Sheth, R.; Mackey, L.; and Fusi, N. 2020.
\newblock Weighted Meta-Learning.
\newblock \emph{arXiv preprint arXiv:2003.09465} .

\bibitem[{Chiuso and Pillonetto(2019)}]{chiuso2019system}
Chiuso, A.; and Pillonetto, G. 2019.
\newblock System identification: A machine learning perspective.
\newblock \emph{Annual Review of Control, Robotics, and Autonomous Systems} 2:
  281--304.

\bibitem[{Cohn, Ghahramani, and Jordan(1996)}]{cohn1996active}
Cohn, D.~A.; Ghahramani, Z.; and Jordan, M.~I. 1996.
\newblock Active learning with statistical models.
\newblock \emph{Journal of artificial intelligence research} 4: 129--145.

\bibitem[{Dai et~al.(2007)Dai, Yang, Xue, and Yu}]{transfer}
Dai, W.; Yang, Q.; Xue, G.-R.; and Yu, Y. 2007.
\newblock Boosting for Transfer Learning.
\newblock In \emph{Proceedings of the 24th International Conference on Machine
  Learning}, ICML ’07, 193–200. New York, NY, USA: Association for
  Computing Machinery.
\newblock ISBN 9781595937933.
\newblock \doi{10.1145/1273496.1273521}.
\newblock \urlprefix\url{https://doi.org/10.1145/1273496.1273521}.

\bibitem[{De~Farias and Van~Roy(2003)}]{de2003linear}
De~Farias, D.~P.; and Van~Roy, B. 2003.
\newblock The linear programming approach to approximate dynamic programming.
\newblock \emph{Operations research} 51(6): 850--865.

\bibitem[{{Du} et~al.(2017){Du}, {El-Khamy}, {Lee}, and {Davis}}]{7926694}
{Du}, X.; {El-Khamy}, M.; {Lee}, J.; and {Davis}, L. 2017.
\newblock Fused DNN: A Deep Neural Network Fusion Approach to Fast and Robust
  Pedestrian Detection.
\newblock In \emph{2017 IEEE Winter Conference on Applications of Computer
  Vision (WACV)}, 953--961.

\bibitem[{Fallah, Mokhtari, and Ozdaglar(2020)}]{fallah2020convergence}
Fallah, A.; Mokhtari, A.; and Ozdaglar, A. 2020.
\newblock On the convergence theory of gradient-based model-agnostic
  meta-learning algorithms.
\newblock In \emph{International Conference on Artificial Intelligence and
  Statistics}, 1082--1092.

\bibitem[{Finn, Abbeel, and Levine(2017)}]{finn2017model}
Finn, C.; Abbeel, P.; and Levine, S. 2017.
\newblock Model-agnostic meta-learning for fast adaptation of deep networks.
\newblock In \emph{Proceedings of the 34th International Conference on Machine
  Learning-Volume 70}, 1126--1135.

\bibitem[{Franceschi et~al.(2018)Franceschi, Frasconi, Salzo, Grazzi, and
  Pontil}]{hyper}
Franceschi, L.; Frasconi, P.; Salzo, S.; Grazzi, R.; and Pontil, M. 2018.
\newblock Bilevel Programming for Hyperparameter Optimization and
  Meta-Learning.
\newblock \emph{ICML 2018} \doi{https://arxiv.org/abs/1806.04910}.

\bibitem[{Gittens and Dempster(1979)}]{gittins-theory}
Gittens, J.; and Dempster, M. 1979.
\newblock Bandit Processes and Dynamic Allocation Indices [with discussion].
\newblock \emph{Journal of the Royal Statistical Society. Series B:
  Methodological} 41: 148--177.
\newblock \doi{10.1111/j.2517-6161.1979.tb01068.x}.

\bibitem[{Gittins, Glazebrook, and Weber(2011)}]{gittins2011multi}
Gittins, J.; Glazebrook, K.; and Weber, R. 2011.
\newblock \emph{Multi-armed bandit allocation indices}.
\newblock John Wiley \& Sons.

\bibitem[{Gittins(1979)}]{gittins1979bandit}
Gittins, J.~C. 1979.
\newblock Bandit processes and dynamic allocation indices.
\newblock \emph{Journal of the Royal Statistical Society: Series B
  (Methodological)} 41(2): 148--164.

\bibitem[{Gopalakrishnan et~al.(2017)Gopalakrishnan, Khaitan, Choudhary, and
  Agrawal}]{GOPALAKRISHNAN2017322}
Gopalakrishnan, K.; Khaitan, S.~K.; Choudhary, A.; and Agrawal, A. 2017.
\newblock Deep Convolutional Neural Networks with transfer learning for
  computer vision-based data-driven pavement distress detection.
\newblock \emph{Construction and Building Materials} 157: 322 -- 330.
\newblock ISSN 0950-0618.
\newblock \doi{https://doi.org/10.1016/j.conbuildmat.2017.09.110}.
\newblock
  \urlprefix\url{http://www.sciencedirect.com/science/article/pii/S0950061817319335}.

\bibitem[{Im, Jiang, and Verma(2019)}]{im2019model}
Im, D.~J.; Jiang, Y.; and Verma, N. 2019.
\newblock Model-Agnostic Meta-Learning using Runge-Kutta Methods.
\newblock \emph{arXiv preprint arXiv:1910.07368} .

\bibitem[{Kingma and Ba(2014)}]{kingma2014adam}
Kingma, D.~P.; and Ba, J. 2014.
\newblock Adam: A method for stochastic optimization.
\newblock \emph{arXiv preprint arXiv:1412.6980} .

\bibitem[{Koppel et~al.(2016)Koppel, Fink, Warnell, Stump, and
  Ribeiro}]{koppel2016online}
Koppel, A.; Fink, J.; Warnell, G.; Stump, E.; and Ribeiro, A. 2016.
\newblock Online learning for characterizing unknown environments in ground
  robotic vehicle models.
\newblock In \emph{2016 IEEE/RSJ International Conference on Intelligent Robots
  and Systems (IROS)}, 626--633. IEEE.

\bibitem[{Krishnamurthy(2016)}]{krishnamurthy2016partially}
Krishnamurthy, V. 2016.
\newblock \emph{Partially Observed Markov Decision Processes}.
\newblock Cambridge University Press.

\bibitem[{Krizhevsky(2009)}]{krizhevsky2009learning}
Krizhevsky, A. 2009.
\newblock Learning Multiple Layers of Features from Tiny Images.
\newblock \emph{Master's thesis, University of Tront} .

\bibitem[{Krizhevsky(2012)}]{cifar}
Krizhevsky, A. 2012.
\newblock Learning Multiple Layers of Features from Tiny Images.
\newblock \emph{University of Toronto} .

\bibitem[{Lai and Robbins(1985)}]{regret}
Lai, T.; and Robbins, H. 1985.
\newblock Asymptotically Efficient Adaptive Allocation Rules.
\newblock \emph{Adv. Appl. Math.} 6(1): 4–22.
\newblock ISSN 0196-8858.
\newblock \doi{10.1016/0196-8858(85)90002-8}.
\newblock \urlprefix\url{https://doi.org/10.1016/0196-8858(85)90002-8}.

\bibitem[{Lattimore and Szepesv{\'a}ri(2020)}]{lattimore2020bandit}
Lattimore, T.; and Szepesv{\'a}ri, C. 2020.
\newblock \emph{Bandit algorithms}.
\newblock Cambridge University Press.

\bibitem[{Learned-Miller(2011)}]{miller}
Learned-Miller, E.~G. 2011.
\newblock Supervised Learning and Bayesian Classification
  \urlprefix\url{https://people.cs.umass.edu/~elm/Teaching/Docs/supervised.pdf}.

\bibitem[{LeCun(1998)}]{lecun1998mnist}
LeCun, Y. 1998.
\newblock The MNIST database of handwritten digits.
\newblock \emph{http://yann. lecun. com/exdb/mnist/} .

\bibitem[{{Lecun} et~al.(1998){Lecun}, {Bottou}, {Bengio}, and
  {Haffner}}]{726791}
{Lecun}, Y.; {Bottou}, L.; {Bengio}, Y.; and {Haffner}, P. 1998.
\newblock Gradient-based learning applied to document recognition.
\newblock \emph{Proceedings of the IEEE} 86(11): 2278--2324.

\bibitem[{Likhosherstov et~al.(2020)Likhosherstov, Song, Choromanski, Davis,
  and Weller}]{likhosherstov2020ufo}
Likhosherstov, V.; Song, X.; Choromanski, K.; Davis, J.; and Weller, A. 2020.
\newblock UFO-BLO: Unbiased First-Order Bilevel Optimization.
\newblock \emph{arXiv preprint arXiv:2006.03631} .

\bibitem[{Liu and Vicente(2019)}]{liu2019stochastic}
Liu, S.; and Vicente, L.~N. 2019.
\newblock The stochastic multi-gradient algorithm for multi-objective
  optimization and its application to supervised machine learning.
\newblock \emph{arXiv preprint arXiv:1907.04472} .

\bibitem[{Liu et~al.(2016)Liu, Racah, Prabhat, Correa, Khosrowshahi, Lavers,
  Kunkel, Wehner, and Collins}]{DBLP:journals/corr/LiuRPCKLKWC16}
Liu, Y.; Racah, E.; Prabhat; Correa, J.; Khosrowshahi, A.; Lavers, D.; Kunkel,
  K.; Wehner, M.~F.; and Collins, W.~D. 2016.
\newblock Application of Deep Convolutional Neural Networks for Detecting
  Extreme Weather in Climate Datasets.
\newblock \emph{CoRR} abs/1605.01156.
\newblock \urlprefix\url{http://arxiv.org/abs/1605.01156}.

\bibitem[{Murphy(2012)}]{murphy2012machine}
Murphy, K.~P. 2012.
\newblock \emph{Machine learning: a probabilistic perspective}.

\bibitem[{Nicholas et~al.(2020)Nicholas, Kuo, Harandi, Fourrier, Walder,
  Ferraro, and Suominen}]{nicholas2020m2sgd}
Nicholas, I.; Kuo, H.; Harandi, M.; Fourrier, N.; Walder, C.; Ferraro, G.; and
  Suominen, H. 2020.
\newblock M2SGD: Learning to Learn Important Weights.
\newblock In \emph{2020 IEEE/CVF Conference on Computer Vision and Pattern
  Recognition Workshops (CVPRW)}, 957--964. IEEE Computer Society.

\bibitem[{Nocedal and Wright(2006)}]{nocedal2006numerical}
Nocedal, J.; and Wright, S. 2006.
\newblock \emph{Numerical optimization}.
\newblock Springer Science \& Business Media.

\bibitem[{{Pan} et~al.(2012){Pan}, {Liu}, {Wang}, {Hu}, and {Jiang}}]{6423452}
{Pan}, J.; {Liu}, C.; {Wang}, Z.; {Hu}, Y.; and {Jiang}, H. 2012.
\newblock Investigation of deep neural networks (DNN) for large vocabulary
  continuous speech recognition: Why DNN surpasses GMMS in acoustic modeling.
\newblock In \emph{2012 8th International Symposium on Chinese Spoken Language
  Processing}, 301--305.

\bibitem[{Park and Oliva(2019)}]{park2019meta}
Park, E.; and Oliva, J.~B. 2019.
\newblock Meta-curvature.
\newblock In \emph{Advances in Neural Information Processing Systems},
  3314--3324.

\bibitem[{Peters, Ruder, and Smith(2019)}]{peters2019tune}
Peters, M.~E.; Ruder, S.; and Smith, N.~A. 2019.
\newblock To Tune or Not to Tune? Adapting Pretrained Representations to
  Diverse Tasks.
\newblock \emph{ACL 2019} 7.

\bibitem[{Puterman(2014)}]{puterman2014markov}
Puterman, M.~L. 2014.
\newblock \emph{Markov decision processes: discrete stochastic dynamic
  programming}.
\newblock John Wiley \& Sons.

\bibitem[{Racah et~al.(2017)Racah, Beckham, Maharaj, Kahou, Prabhat, and
  Pal}]{NIPS2017_6932}
Racah, E.; Beckham, C.; Maharaj, T.; Kahou, S.; Prabhat, M.; and Pal, C. 2017.
\newblock ExtremeWeather: A large-scale climate dataset for semi-supervised
  detection, localization, and understanding of extreme weather events.
\newblock In Guyon, I.; Luxburg, U.~V.; Bengio, S.; Wallach, H.; Fergus, R.;
  Vishwanathan, S.; and Garnett, R., eds., \emph{Advances in Neural Information
  Processing Systems 30}, 3405--3416. Curran Associates, Inc.
\newblock
  \urlprefix\url{http://papers.nips.cc/paper/6932-extremeweather-a-large-scale-climate-dataset-for-semi-supervised-detection-localization-and-understanding-of-extreme-weather-events.pdf}.

\bibitem[{Radford, Metz, and Chintala(2015)}]{radford2015unsupervised}
Radford, A.; Metz, L.; and Chintala, S. 2015.
\newblock Unsupervised Representation Learning with Deep Convolutional
  Generative Adversarial Networks.

\bibitem[{Rajeswaran et~al.(2019)Rajeswaran, Finn, Kakade, and
  Levine}]{rajeswaran2019metalearning}
Rajeswaran, A.; Finn, C.; Kakade, S.; and Levine, S. 2019.
\newblock Meta-Learning with Implicit Gradients.

\bibitem[{Settles(2011)}]{settles2011theories}
Settles, B. 2011.
\newblock From theories to queries: Active learning in practice.
\newblock In \emph{Active Learning and Experimental Design workshop In
  conjunction with AISTATS 2010}, 1--18.

\bibitem[{Shorten and Khoshgoftaar(2019)}]{shorten2019survey}
Shorten, C.; and Khoshgoftaar, T.~M. 2019.
\newblock A survey on image data augmentation for deep learning.
\newblock \emph{Journal of Big Data} 6(1): 60.

\bibitem[{Simon et~al.(2020)Simon, Koniusz, Nock, and
  Harandi}]{simon2020modulating}
Simon, C.; Koniusz, P.; Nock, R.; and Harandi, M. 2020.
\newblock On modulating the gradient for meta-learning.
\newblock ECCV.

\bibitem[{Song et~al.(2019)Song, Gao, Yang, Choromanski, Pacchiano, and
  Tang}]{song2019maml}
Song, X.; Gao, W.; Yang, Y.; Choromanski, K.; Pacchiano, A.; and Tang, Y. 2019.
\newblock ES-MAML: Simple Hessian-Free Meta Learning.
\newblock In \emph{International Conference on Learning Representations}.

\bibitem[{Tan et~al.(2018)Tan, Sun, Kong, Zhang, Yang, and
  Liu}]{10.1007/978-3-030-01424-7_27}
Tan, C.; Sun, F.; Kong, T.; Zhang, W.; Yang, C.; and Liu, C. 2018.
\newblock A Survey on Deep Transfer Learning.
\newblock In K{\r{u}}rkov{\'a}, V.; Manolopoulos, Y.; Hammer, B.; Iliadis, L.;
  and Maglogiannis, I., eds., \emph{Artificial Neural Networks and Machine
  Learning -- ICANN 2018}, 270--279. Cham: Springer International Publishing.
\newblock ISBN 978-3-030-01424-7.

\bibitem[{{Varaiya}, {Walrand}, and {Buyukkoc}(1985)}]{1103989}
{Varaiya}, P.; {Walrand}, J.; and {Buyukkoc}, C. 1985.
\newblock Extensions of the multiarmed bandit problem: The discounted case.
\newblock \emph{IEEE Transactions on Automatic Control} 30(5): 426--439.

\bibitem[{Wang et~al.(2019)Wang, Lehman, Clune, and Stanley}]{wang2019paired}
Wang, R.; Lehman, J.; Clune, J.; and Stanley, K.~O. 2019.
\newblock Paired open-ended trailblazer (poet): Endlessly generating
  increasingly complex and diverse learning environments and their solutions.
\newblock \emph{arXiv preprint arXiv:1901.01753} .

\bibitem[{Wang et~al.(2018)Wang, Wu, Li, Gu, Xiang, Zhang, and
  Li}]{wang2018large}
Wang, Y.; Wu, X.-M.; Li, Q.; Gu, J.; Xiang, W.; Zhang, L.; and Li, V.~O. 2018.
\newblock Large Margin Meta-Learning for Few-Shot Classification.
\newblock In \emph{Neural Information Processing Systems (NIPS) Workshop on
  Meta-Learning, Montreal, Canada}.

\bibitem[{Xu, Krzyzak, and Suen(1992)}]{xu1992methods}
Xu, L.; Krzyzak, A.; and Suen, C.~Y. 1992.
\newblock Methods of combining multiple classifiers and their applications to
  handwriting recognition.
\newblock \emph{IEEE transactions on systems, man, and cybernetics} 22(3):
  418--435.

\bibitem[{Yin et~al.(2019)Yin, Tucker, Zhou, Levine, and
  Finn}]{yin2019metalearning}
Yin, M.; Tucker, G.; Zhou, M.; Levine, S.; and Finn, C. 2019.
\newblock Meta-Learning without Memorization.

\bibitem[{Yin et~al.(2017)Yin, Kann, Yu, and Schütze}]{yin2017comparative}
Yin, W.; Kann, K.; Yu, M.; and Schütze, H. 2017.
\newblock Comparative Study of CNN and RNN for Natural Language Processing.

\bibitem[{Yu et~al.(2018)Yu, Finn, Xie, Dasari, Zhang, Abbeel, and
  Levine}]{yu2018one}
Yu, T.; Finn, C.; Xie, A.; Dasari, S.; Zhang, T.; Abbeel, P.; and Levine, S.
  2018.
\newblock One-shot imitation from observing humans via domain-adaptive
  meta-learning.
\newblock \emph{arXiv preprint arXiv:1802.01557} .

\end{thebibliography}
